# Consecutive Support: Better Be Close!*

Edgar de Graaf, Jeannette de Graaf, and Walter A. Kosters

Leiden Institute of Advanced Computer Science, Leiden University, The Netherlands
{edegraaf,graaf,kosters}@liacs.nl

**Abstract.** We propose a new measure of support (the number of occurrences of a pattern), in which instances are more important if they occur with a certain frequency and close after each other in the stream of transactions. We will explain this new consecutive support and discuss how patterns can be found faster by pruning the search space, for instance using so-called parent support recalculation. Both consecutiveness and the notion of hypercliques are incorporated into the ECLAT algorithm. Synthetic examples show how interesting phenomena can now be discovered in the datasets. The new measure can be applied in many areas, ranging from bio-informatics to trade, supermarkets, and even law enforcement. E.g., in bio-informatics it is important to find patterns contained in many individuals, where patterns close together in one chromosome are more significant.

## 1 Introduction

In earlier research we explored the use of frequent itemsets to visualize deviations in chromosome data concerning people with a certain illness, genomic profiling [4]. During our exploration of this problem it became apparent that patterns are more important when the areas (transactions) in which they occur are close together. The consecutiveness of transactions containing the pattern plays an important role in some applications. Patterns are frequent sets of items where frequent means that their support, that can be defined in different ways, is more than the *minsup* threshold. In the biological problem the items are individuals and the transactions are "clones", pieces of the chromosome that might occur more or less often than in a healthy individual. Patterns in close transactions are better because they are close together in the chromosome and are biologically more significant than patterns that are far apart and in different chromosomes.

*Consecutive support* informally is the support or the number of occurrences of patterns where we take into account the distance between transactions where the pattern occurred. With distance is meant the number of in-between transactions that did not contain the pattern. Of course, this only makes sense if the transactions are given in some logical order. This type of support can be applicable in a number of domains:

---

* This research is carried out within the Netherlands Organization for Scientific Research (NWO) MISTA Project (grant no. 612.066.304).

- **Supermarket**. E.g., big supermarkets receive large quantities of goods every day. Knowing which goods will be sold in large quantities close in time helps the supermarket decide when to refill these goods.
- **Trading**. E.g., a combination of stock being sold once may lead to waves of these stocks being sold close after each other while other combinations might not.
- **Law enforcement**. E.g., when police investigates telephone calls, subjects that are discussed during a longer period might be more interesting than subjects (word combinations) that are mentioned often at separate moments.

This type of support still needs to be defined and its usefulness needs to be shown. To this end, this paper makes the following contributions:

**We define different variations of consecutive support, having two parameters.** We will define this support with a reward factor $\rho$ and a punishment factor $\sigma$. And we will show how this can be implemented.

**We will show how to speed up the search by pruning the search space.** Some methods of pruning will not give all patterns, but we can find most of the important patterns faster. Other pruning methods will not influence the outcome, but they require more calculation and the speed up is less.

**We will show the usefulness of consecutive support using a motivating example.** With our experimental results we show how consecutive support, compared to the results in [4], gives new and interesting patterns when applied to the biological problem of finding patterns in chromosomes.

This research is related to work done on the (re)definition of support, gap constraint and weighted association rule mining. The notion of support was first introduced by Agrawal et al. in [1] in 1993. Much later Steinbach et al. in [7] generalized the notion of support providing a framework for different definitions of support in the future. Our notion of consecutive support is not easily fitted in the eval-function provided there. (Next to this framework Steinbach also provides a couple of example functions.) Our work also has some relation with research done in [8] concerning weighted association rule mining where different items have different weights. Consecutive support can be seen as weighted patterns based on distance between transactions that contain them.

If we take a market basket database as an example, we will have a database where the customers (or transactions) are itemsets of products they bought. We could invert this database such that transactions correspond to the products, and are itemsets of customers that bought the product. Now we can search for patterns and with techniques like the time window constraint as defined in [5] or the gap constraint as defined in [2], we can search for customers who bought products close in time. However the combination of products that were bought will be lost. Furthermore in our case we want to know which products occur often in combinations.

Finally this work is related to some of our earlier work. Primarily the work done in [4] already stated that the biological problem could profit from incorporating consecutiveness into frequent itemset mining. Secondly in [3] it was mentioned that support is just another measure of saying how good a pattern fits with the data. There we defined different variations of this measure, and consecutive support can been seen as such a variation.

The formal definitions concerning consecutive support are given in Section 2. Pruning methods are discussed in Section 3. In Section 4 we present experimental results, and we conclude in Section 5.

## 2 Consecutive Support

The definition of association rules relies on that of support: the number of transactions that contain a given itemset. In this paper we propose a more general definition, that takes the consecutiveness of the transactions into account.

Suppose items are from the set $\mathcal{I} = \{1, 2, \ldots, n\}$, where $n \geq 1$ is a fixed integer constant. A *transaction* is an *itemset*, which is a subset of $\mathcal{I}$. A *database* is an *ordered series* of $m$ transactions, where $m \geq 1$ is a fixed integer constant. If an itemset is an element of a database, it is usually referred to as a transaction.

The *traditional support* of an itemset $I$ with respect to a database $\mathcal{D}$, denoted by $\text{TradSupp}(I, \mathcal{D})$, is the number of transactions from $\mathcal{D}$ that contain $I$. Clearly, $0 \leq \text{TradSupp}(I, \mathcal{D}) \leq m$.

We now propose a more general definition. Fix two real parameters $\rho \geq 0$ and $0 \leq \sigma \leq 1$. Suppose we have an itemset $I$ and let $O_j \in \{0, 1\}$ ($j = 1, 2, \ldots, m$) denote whether or not the $j^{\text{th}}$ transaction in the database $\mathcal{D}$ contains $I$ ($O_j$ is 1 if it does contain $I$, and 0 otherwise; the $O$'s are referred to as the *O-series*). The following algorithm computes a real value $t$ in one linear sweep through the database and the resulting $t$ is defined as the *consecutive support* of $I$ with respect to $\mathcal{D}$ (denoted by $\text{Supp}(I, \mathcal{D}, \rho, \sigma)$):

```
t := 0; j := 1; reward := 0;
while ( j ≤ m ) do
    if ( O_j = 1 ) then
        t := t + 1 + reward; reward := reward + ρ;
    else
        reward := reward · σ;
    fi
    j := j + 1;
od
```

The consecutive support $t$ can become very large, and one could for example use $\sqrt{t}$ instead. In our examples we will not use $\sqrt{t}$, and just employ $t$.

*Example 1.* Assume the O-serie of a certain pattern $I$ equals 101101, $\rho = 1$ and $\sigma = 0.1$. The consecutive support $t$ will then be 5.41:

| $O$ | 1 | 0 | 1 | 1 | 0 | 1 |
|---|---|---|---|---|---|---|
| *reward* | 0 | 1 | 0.1 | 1.1 | 2.1 | 0.21 |
| $t$ | 1 | 1 | 2.1 | 4.2 | 4.2 | **5.41** |

Note that during the loop the value of *reward*, which "rewards" the occurrence of a 1, is always at least 0. If *reward* would never be adapted, i.e., it would remain 0 all the time, this algorithm would compute $\mathrm{TradSupp}(I, \mathcal{D})$. We easily see that $0 \leq \mathrm{Supp}(I, \mathcal{D}, \rho, \sigma) \leq m + m(m-1)\rho/2$. The maximum value is obtained if and only if all transactions from the database $\mathcal{D}$ contain $I$, i.e., an $O$-series entirely consisting of 1's. Only the all 0's series gives the minimum value 0. Furthermore we have for any $0 \leq \sigma \leq 1$: $\mathrm{Supp}(I, \mathcal{D}, 0, \sigma) = \mathrm{TradSupp}(I, \mathcal{D})$. For all $\rho \geq 0$ and $0 \leq \sigma \leq 1$, $\mathrm{Supp}(I, \mathcal{D}, \rho, \sigma) \geq \mathrm{TradSupp}(I, \mathcal{D})$ holds. Finally, note that the so-called APRIORI property [1] or anti-monotonicity constraint is satisfied: for all $\rho \geq 0$ and $0 \leq \sigma \leq 1$, $\mathrm{Supp}(I, \mathcal{D}, \rho, \sigma) \geq \mathrm{Supp}(I', \mathcal{D}, \rho, \sigma)$ if the itemset $I'$ contains the itemset $I$. This follows from the observation that the *reward*-values in the $I'$-case are never larger than those in the $I$-case.

It is not hard to show that for the $O$-series $1^{a_1}0^{b_1}1^{a_2}0^{b_2}\ldots 0^{b_{n-1}}1^{a_n}$ (a series of $a_1$ 1's, $b_1$ 0's, $a_2$ 1's, $b_2$ 0's, ..., $b_{n-1}$ 0's, $a_n$ 1's) consecutive support equals

$$\sum_{i=1}^{n} a_i + \rho \sum_{i=1}^{n} a_i(a_i-1)/2 + \rho \sum_{1 \leq i < j \leq n} a_i a_j \sigma^{b_i+b_{i+1}+\cdots+b_{j-1}} =$$
$$(1-\rho/2)S + \rho S^2/2 - \rho \sum_{1 \leq i < j \leq n} a_i a_j (1 - \sigma^{b_i+b_{i+1}+\cdots+b_{j-1}}),$$

where $S = \sum_{i=1}^{n} a_i$; here $0^0$ must be interpreted as 1 (an exponent 0 can be avoided by demanding all $b_i$'s to be non-zero; if we also demand all $a_i$'s to be $> 0$ both the number $n$ and the numbers $a_i$ and $b_i$ are unique, given an $O$-series). The formula follows from the fact that if *reward* equals $\varepsilon$, then the series $1^k 0^\ell$ changes this into $(\varepsilon + k\rho) \cdot \sigma^\ell$, meanwhile giving a contribution of $k + k\varepsilon + k(k-1)\rho/2$ to the consecutive support. An extra series $0^\ell$ at the beginning or end has no influence on the consecutive support.

The second part of the equation $\rho \sum_{i=1}^{n} a_i(a_i-1)/2$, consist of the $\rho$'s added for a subset of consecutive 1's in the $O$-serie. The last part of the equation is the addition of the rewards from the previous consecutive 1's decreased with $\sigma$ because of the number of 0's between the groups of consecutive 1's. Also note that when we choose $\rho = 2$ we get $S^2 - \rho \sum_{1 \leq i < j \leq n} a_i a_j (1 - \sigma^{b_i+b_{i+1}+\cdots+b_{j-1}})$. This shows that consecutive support is at most $S^2$ if $\rho = 2$.

*Example 2.* Take $\rho = 2$. Then the $O$-series $1^5 0^\ell 1^4$ has consecutive support $81 - 40(1 - \sigma^\ell)$. As $\ell \to \infty$ this value approaches $41 = 5^2 + 4^2$, whereas for small $\ell$ and $\sigma \approx 1$ it is near $81 = (5+4)^2$.

It can be observed that the consecutive support as defined above only depends on the lengths of the "runs" and the lengths of the intermediate "non-runs": the $a_i$'s and $b_i$'s above. Here a *run* is defined as a maximal consecutive series of 1's in a 0/1 sequence. Indeed, the sum $\sum_{k=i}^{j-1} b_k$ equals the number of 0's between run $i$ and run $j$. This also implies that the definition is *symmetric*, in the sense that the support is unchanged if the order of the $O$-series is reversed — a property that is certainly required. (In fact, this is due to the fact that $\rho$ is added, while we multiply by $\sigma$.)

Instead of this way of calculating consecutive support it is also possible to augment the *O*-serie with *time stamps*. Then one is able to use the real time between two transactions in calculating the consecutive support. In the previous definition each transaction was assumed to take the same amount of time. Another improvement might be to reinitialize *reward* to 0 at suitable moments, for instance at chromosome boundaries or at "closing hours".

We now consider algorithms that find all frequent itemsets, given a database. A *frequent* itemset is an itemset with support at least equal to some pre-given threshold, the so-called *minsup*. Thanks to the APRIORI property many efficient algorithms exist. However, the really fast ones rely upon the concept of FP-TREE or something similar, which does not keep track of consecutivity. This makes these algorithms hard to adapt for consecutive support.

One fast algorithm that does not make use of FP-TREES is called ECLAT [10]. ECLAT grows patterns recursively while remembering which transactions contained the pattern, making it very suitable for consecutive support. In the next recursive step only these transactions are considered when counting the occurrence of a pattern. All counting is done by using a matrix and patterns are extended with new items using the order in the matrix. This can easily be adapted to incorporate consecutiveness.

## 3 Pruning Methods

The consecutive support of patterns can be much higher than the traditional support. As a consequence more patterns will be frequent or the minimal support threshold should be set much higher. So it is important to use pruning. In this work we propose several pruning methods. These are implemented in our version of ECLAT, which counts consecutive support, from here on called CONSECLAT. Some pruning methods influence the completeness: you will not get all patterns.

### 3.1 Parent Support Recalculation

The first pruning method we discuss does not affect completeness. Basically this *parent support recalculation* method does the following for each transaction $r$:

- Calculate the consecutive support the parent had collected when considering transaction $r$, where the *child* is the current itemset being the *parent* itemset generated in the previous recursive step extended with one item.
- Subtract this support from the total support of the parent.
- Add to this the support the child pattern has collected up until now. The child can still maximally achieve this consecutive support, from here on called *maximal achievable support*.
- Return a support of 0 if this is less then the minimal support.

In re-calculating the support of the parent pattern at a certain transaction we make use of the fact that we store which transactions contained the parent pattern. In CONSECLAT we use a list of transaction numbers that contain the pattern. With these numbers we can (re)calculate the consecutive support of the parent in the same loop through the database:

$$reward\_parent := reward\_parent \cdot \sigma^{diff}$$
$$partial\_support\_parent := partial\_support\_parent + 1 + reward\_parent$$
$$reward\_parent := reward\_parent + \rho$$

where *diff* is the number of transactions that did not contain the parent pattern:

$$diff := current\_transaction\_number - last\_transaction\_number - 1$$

Here *last_transaction_number* is the transaction number of the last transaction (before *current_transaction_number*, the current one) that contained the parent pattern. Now the maximal achievable support for the child can be calculated:

$$possible := parent\_total - partial\_support\_parent + support$$

The variable *parent_total* is the support the parent pattern was able to achieve and *support* is the consecutive support that the child-pattern was able to "collect" until the current transaction.

Now the algorithm will stop counting support if it is impossible to still achieve a support that is higher than the minimal support. The child pattern can at most get the maximal achievable support, because it can never score better than its parent on the remaining transactions.

*Example 3.* Assume the following "child"-pattern that is an extension of the "parent"-pattern:

| $O_{parent}$ | 1 | 1 | 1 | 0 | 0 | 1 |
|---|---|---|---|---|---|---|
| $O_{child}$ | 0 | 0 | 0 | 0 | 0 | 1 |

Furthermore assume $minsup = 5, \sigma = 0.1$ and $\rho = 1.0$; then we can stop counting support when we encounter the second zero. At that point we know that at most we can get a consecutive support of 4.03 (the consecutive support of the parent was 7.03 and a consecutive support of 3 was lost in the child).

### 3.2 Introducing $\alpha$

In the parent support recalculation method we assumed that the remaining transactions will all contain the child. This is an optimistic estimate necessary for guaranteeing completeness. We could assume that the child pattern will be contained in less than all of the remaining transactions by a factor $\alpha$, $0 \leq \alpha \leq 1$. We then introduce this $\alpha$ in the calculation of maximal achievable support:

$$possible := \alpha \cdot (parent\_total - partial\_support\_parent) + support$$

This will speed up the mining process, but we lose completeness.

### 3.3 Exact Depth

In the case of our motivating example biologists wanted to visualize only long patterns, because the small patterns are so numerous that affected areas are less recognizable. This wish to only get patterns of a certain length can be used for pruning. We allow the user to set the maximal length $\eta$ that patterns should have. Now we can prune if the following holds:

$$last\_frequent\_item - item < \eta - depth$$

Here items are numbers lexicographically ordered in the matrix used by CON-SECLAT, *last_frequent_item* is the last item in that matrix that is still frequent and *item* is the current item that we are considering. The *depth* is the recursive depth, which is equal to the length of the pattern. If the above statement holds then the pattern will never reach the required length $\eta$ and it can be pruned.

*Example 4.* Assume the following database matrix:

| item numbers: | 1 | 2 | 3 | 4 |
|---|---|---|---|---|
| transaction 1 | 1 | 0 | 1 | 1 |
| transaction 2 | 1 | 0 | 0 | 1 |
| transaction 3 | 0 | 1 | 1 | 1 |

Say $\eta = 4$, the parent item set is $\{1\}$ and we are considering to extend this with $\{3\}$ to the child $\{1,3\}$. However $4 - 3 < 4 - 2$ and so $\{1,3\}$ and all its children are pruned.

### 3.4 Hyperclique Patterns and *h*-confidence

Many principles applicable to traditional support can still be used when one considers consecutive support. In the case of our working example we wanted to visualize patterns with a minimal consecutive support of 25. Unfortunately there are many patterns with this support. In order to speed up the search and to filter out uninteresting patterns we can search for *hyperclique patterns* as described in [9]. Because of space limitations we explain hyperclique patterns via an example:

*Example 5.* First a *minimal confidence threshold* $h_c$ is defined, say $h_c = 0.6$. We want to know if $\{A, B, C\}$ is a hyperclique pattern. We calculate the confidence of $A \to \{B, C\}$, $B \to \{A, C\}$ and $C \to \{A, B\}$. The lowest of these confidences is the *h-confidence*, which must be higher then $h_c$. Assume that $conf(A \to B, C) = \text{Supp}(\{A, B, C\}, \mathcal{D}, \rho, \sigma) / \text{Supp}(\{A\}, \mathcal{D}, \rho, \sigma) = 0.58$. Then $\{A, B, C\}$ is no hyperclique pattern.

When we combine the concept of consecutive support with hyperclique patterns we get patterns that occur frequent but in the flow of transactions close after each other and there is a *strong affinity* between items: the presence of $x \in P$, where $P$ is an item set, in a transaction strongly implies the presence of the other items in $P$.

Hyperclique patterns possess the *cross-support property*. This means that we will not get *cross-support patterns*. These are patterns containing items of substantially different support levels.

We can easily see that hyperclique patterns possess the cross-support property. If one item has a high support and another item has a low support then $h$-confidence will be low if the denominator is the item with the high support.

*Example 6.* Say $A$ is an item with a consecutive support of 200 and $B$ has a consecutive support 50. The support of $\{A, B, C\}$ will at most be 50 because of the APRIORI property (the support of the superset is always the same or less than the support of its subsets). So the confidence of $conf(A \to B, C)$ can at most be $50/200 = 0.25$. As a consequence the $h$-confidence of $\{A, B, C\}$ will also be at most 0.25. And if $h_c = 0.6$ then $\{A, B, C\}$ and all the patterns that are grown from it can be pruned.

The combination of hyperclique patterns and consecutive support allows us to find patterns that occur in transactions that follow each other close, yet minimal support can be relatively low. This property is especially handy for our motivating example, because a minimal consecutive support of 25 will generate many cross-support patterns, which are pruned if we search only for hyperclique patterns. Hyperclique patterns also posses the anti-monotone property, because as patterns grow the numerator of the confidence calculation stays the same or declines. The denominator stays fixed and so $h$-confidence will decrease or stay the same:

*Example 7.* Say $conf(A \to B, C) = 0.58$. The superset $\{A, B, C, D\}$ will at most have the same consecutive support as $\{A, B, C\}$. Also the denominator $\text{Supp}(\{A\}, \mathcal{D}, \rho, \sigma)$ stays the same, so the $h$-confidence of $\{A, B, C, D\}$ can at most be 0.58.

## 4  Results and Performance

The experiments were done for three main reasons. First of all we want to show that *consecutive support can enable one to find new patterns* that one does not find with the traditional support. Secondly we want to show how *using the principle of h-confidence one can filter the data*. Finally we want to show *how pruning speeds up the search* for consecutive patterns.

All experiments were done on a Pentium 4 2.8 GHz with 512MB RAM. For our experiments we used three datasets. One biological dataset, referred to as the *Nakao dataset*, was also used in [4]. This data set originates from Nakao et al. who used the dataset in [6]. This publicly available dataset contains normalized $\log_2$-ratios for 2124 clones, located on chromosomes 1–22 and the X-chromosome. Each clone is a transaction with 2 to 1020 real numbers corresponding to patients. We can look at gains and/or losses. If we consider gains, a patient is present in a transaction (clone) if his value deviates from a healthy person more than 0.225. If we would like to visualize losses then a patient will be in a transaction if its value for this clone is lower than $-0.225$. The other datasets are synthetic datasets made to show how consecutive support can be used to find patterns that could not be found before. The first synthetic data set, referred to as the *food+drink dataset*, describes a cafe-restaurant where in the middle of a day a lot of people buy bread and orange juice; it has 1000 transactions and 100 items. The second synthetic data set will be explained later.

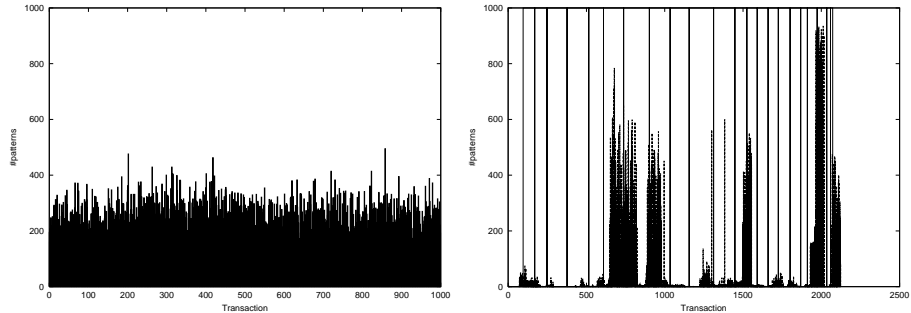

**Fig. 1.** Occurrence graph of food+drink using traditional support ($minsup = 257$)

**Fig. 2.** Occurrence graph of Nakao (gains) using traditional support ($minsup = 129$)

In the experiments of Figure 1, 2, 3 and 4 we tried to find approximately 1000 patterns with the highest traditional or consecutive support. After this we count for each transaction how many patterns it contains, allowing us to see how active areas are. For the Nakao dataset more active means that many clones (gains) in the same area are present in many groups of patients.

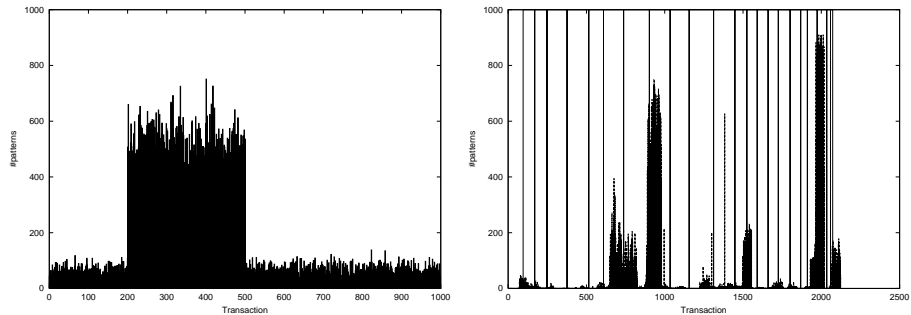

**Fig. 3.** Occurrence graph of food+drink using consecutive support ($minsup = 467$, $\rho = 1.0$ and $\sigma = 0.5$)

**Fig. 4.** Occurrence graph of Nakao (gains) using consecutive support ($minsup = 827$, $\rho = 1.0$ and $\sigma = 0.5$)

Figure 1 and Figure 2 show where patterns occur when we use traditional support, giving results similar to those in [4]. For each transaction the number of patterns that it occurs in is plotted in a so-called *occurrence graph*. In each of these graphs we will indicate chromosome borders when the Nakao dataset is visualized. In the food+drink dataset it is very clear that consecutive support enables us to see new patterns. Figure 3 shows that in certain areas patterns are more consecutive. Figure 4 shows that certain areas are less active if we use consecutive support instead of traditional support and some areas contain more patterns, hence providing patterns that occur together in one part of the chromosome instead of far apart.

In the following experiments the goal was to show that combining hyperclique patterns with consecutive support enables us to see patterns occurring in bursts. In order to show this we created a new synthetic dataset, referred to as the *coffee+cookie dataset*, where in the cafe-restaurant small bursts of people buy coffee and a cookie, during the day in the coffee breaks.

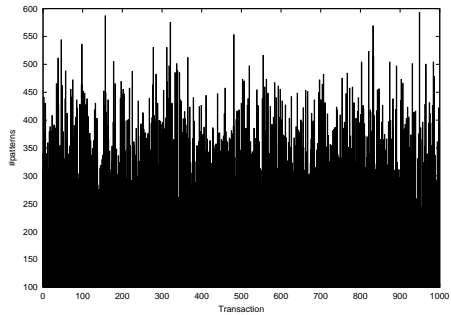
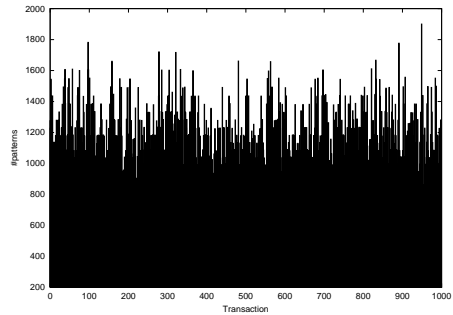

**Fig. 5.** Occurrence graph of coffee+cookie using only $h$-confidence ($minsup = 64$ and $h_c = 0.5$)

**Fig. 6.** Occurrence graph of coffee+cookie using only consecutive support ($minsup = 225$, $\rho = 1.0$, $\sigma = 0.5$ and $h_c = 0$)

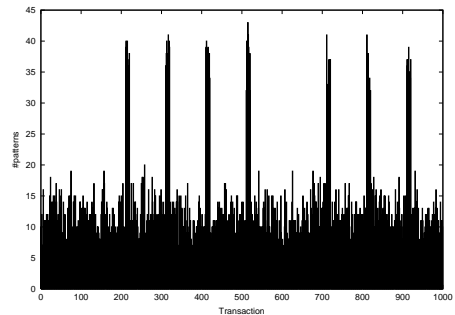

**Fig. 7.** Occurrence graph of coffee+cookie using both consecutive support and $h$-confidence ($minsup = 64$, $\rho = 1.0$, $\sigma = 0.5$ and $h_c = 0.31$)

Figure 5 does not clearly show the small groups buying the same products: just hyperclique patterns do not reveal the bursts. Figure 6 shows that with only consecutive support we are also unable to discover these patterns. Figure 7 shows people buying the products in bursts. Consecutive support stresses patterns that are consecutive and the principle of $h$-confidence filters out the noise caused by cross-support patterns.

When we apply these techniques to the Nakao dataset (losses), in Figure 9, we can see, e.g., on chromosomes 14 and 15 (near transaction 1600) that certain areas become more active compared to not using $h$-confidence in Figure 8.

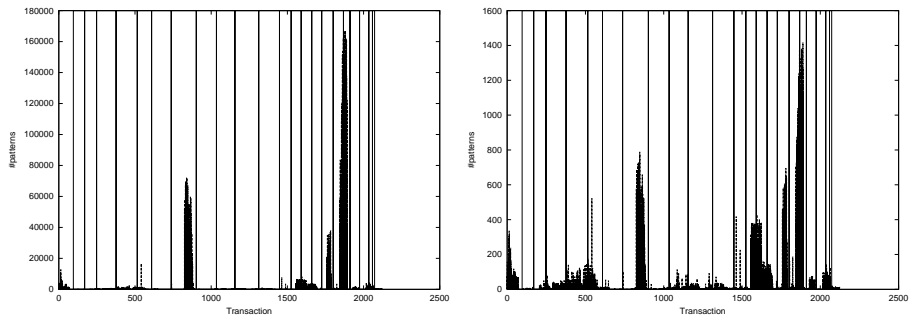

**Fig. 8.** Occurrence graph of Nakao (losses) using only consecutive support ($minsup = 400$, $\rho = 1.0$, $\sigma = 0.9$ and $h_c = 0$)

**Fig. 9.** Occurrence graph of Nakao (losses) using both consecutive support and $h$-confidence ($minsup = 25$, $\rho = 1.0$, $\sigma = 0.9$ and $h_c = 0.15$)

| $\rho = 1.0$, $\sigma = 0.5$ | Nakao (gains) | food+drink | coffee+cookie |
|---|---|---|---|
| Normal | 80 | 263 | 182 |
| With MAS | 44 | 226 | 192 |
| MAS with $\alpha = 0.9$ | 26 | 158 | 144 |
| Exact Depth | 36 | 172 | 149 |

**Table 1.** Time in seconds needed to mine each of the three datasets

We now consider run times where $\rho = 1.0$ and $\sigma = 0.5$. In Table 1 "*Normal*" is ConseClat without any pruning, "*With MAS*" employs the recalculation of parent support in ConseClat to find the maximal achievable support, and "*MAS with $\alpha = 0.9$*" introduces $\alpha = 0.9$. Finally "*Exact Depth*" is almost equal to "*With MAS*", however with the addition that it discovers only the patterns of length $\eta = 10$ and this is used to prune the search space. The table displays the time (in seconds) it took to mine the datasets. The *minsup* of the Nakao dataset (gains) is set to 400. For both synthetic datasets *minsup* is set to 100. In most cases "*With MAS*" outperforms "*Normal*", except for the last dataset. The time saved using this technique is too small compared to the amount of calculation it requires in the case of this dataset. All other results are as expected: using $\alpha$ and "*Exact Depth*" both speed up the search, but they will not give all the patterns "*Normal*" ConseClat gives. E.g., $\alpha = 0.9$ gives 77% of the patterns.

## 5 Conclusions and Future Work

Consecutive support enables us to find new and useful patterns compared to traditional methods. Principles applicable to traditional support can still be used with consecutive support. For instance the combination of consecutive support and the $h$-confidence threshold enables us to find small bursts of patterns. In this case $h$-confidence filters out noise and consecutive support amplifies the bursts.

Consecutive support might result in many more patterns. Because of this, pruning the search space is important. In this paper we proposed a number of methods for pruning, where some methods do not give all patterns.

Using the distance between transactions like it is done in this paper is an interesting area of research. In the future we want to examine if consecutive support enables us to visualize even more types of behavior. Also we want to see if we can speed up the search for consecutive patterns even more. Finally we want to extend consecutive support by using distance between transactions in different ways, which hopefully give us more relevant patterns.

**Acknowledgments** We would like to thank Joost Broekens, Joost Kok, Siegfried Nijssen and Wim Pijls.

## References

1. Agrawal, R., Imielinski, T., Srikant, R.: *Mining Association Rules between Sets of Items in Large Databases*. In Proc. of ACM SIGMOD Conference on Management of Data (1993), pp. 207–216.
2. Antunes, C., Oliveira, A.L.: *Generalization of Pattern-Growth Methods for Sequential Pattern Mining with Gap Constraints*. In Machine Learning and Data Mining in Pattern Recognition (MLDM 2003), LNCS 2734, Springer, pp. 239–251.
3. Graaf, E.H. de, Kosters, W.A.: *Using a Probable Time Window for Efficient Pattern Mining in a Receptor Database*. In Proc. of 3rd Int. ECML/PKDD Workshop on Mining Graphs, Trees and Sequences (MGTS'05), pp. 13–24.
4. Graaf, J.M. de, Menezes, R.X. de, Boer, J.M., Kosters, W.A.: *Frequent Itemsets for Genomic Profiling*. In Proc. 1st International Symposium on Computational Life Sciences (CompLife 2005), LNCS 3695, Springer, pp. 104–116.
5. Leleu, M., Rigotti, C., Boulicaut, J.F., Euvrard, G.: *Constraint-Based Mining of Sequential Patterns over Datasets with Consecutive Repetitions*. In Proc. 7th European Conference on Principles of Data Mining and Knowledge Discovery (PKDD 2003), LNAI 2838, Springer, pp. 303–314
6. Nakao, K., Mehta, K.R., Fridlyand, J., Moore, D.H., Jain, A.N., Lafuente, A., Wiencke, J.W., Terdiman, J.P., Waldman, F.M.: *High-Resolution Analysis of DNA Copy Number Alterations in Colorectal Cancer by Array-Based Comparative Genomic Hybridization*. Carcinogenesis 25 (2004), pp. 1345–1357
7. Steinbach, M., Tan, P., Xiong, H., Kumar, V.: *Generalizing the Notion of Support*. In Proc. 10th Int. Conf. on Knowledge Discovery and Data Mining (KDD '04), pp. 689–694.
8. Tao, F., Murtagh, F., Farid, M.: *Weighted Association Rule Mining using Weighted Support and Significance Framework*. In Proc. 9th Int. Conf. on Knowledge Discovery and Data Mining (KDD '03), pp. 661–666.

9. Xiong, H., Tan, P., Kumar, V.: *Mining Strong Affinity Association Patterns in Data Sets with Skewed Support Distribution*. In Proc. Int. Conf. on Data Mining (ICDM'03), pp. 387–394.
10. Zaki, M., Parthasarathy, S., Ogihara, M., Li, W.: *New Algorithms for Fast Discovery of Association Rules*. In Proc. 3rd Int. Conf. on Knowledge Discovery and Data Mining (KDD '97), pp. 283–296.